\documentclass[letterpaper, 10 pt, conference]{ieeeconf}  

\IEEEoverridecommandlockouts                              

\usepackage{graphics} 
\usepackage{epsfig} 
\usepackage{mathptmx} 
\usepackage{times} 
\usepackage{hyperref}
\usepackage{cite}
\usepackage{amsmath,amssymb,amsfonts}
\usepackage{algorithmic}
\usepackage{graphicx}
\usepackage{textcomp}
\usepackage{xcolor}
\usepackage{adjustbox}
\usepackage{xspace}
\usepackage{multirow}
\usepackage{authblk}
\usepackage[symbol]{footmisc}

\makeatletter
\DeclareRobustCommand\onedot{\futurelet\@let@token\@onedot}
\def\@onedot{\ifx\@let@token.\else.\null\fi\xspace}

\def\eg{\emph{e.g}\onedot}

\def\etal{\emph{et al}\onedot}
\makeatother
\makeatletter
\let\NAT@parse\undefined
\makeatother

\title{
\LARGE \bf
DocDeshadower: Frequency-Aware Transformer for Document Shadow Removal
}

\author[1*]{Ziyang Zhou}
\author[2*]{Yingtie Lei\thanks{* These authors contribute equally to this work.}}
\author[12$^\dag$]{Xuhang Chen}
\author[2]{Shenghong Luo}
\author[3]{Wenjun Zhang}
\author[2$^\dag$]{\authorcr Chi-Man Pun\thanks{$^\dag$ Corresponding author.}}
\author[1]{Zhen Wang}
\affil[1]{School of Computer Science and Engineering, Huizhou University, Huizhou, China}
\affil[2]{Faculty of Science and Technology, University of Macau, Macau, China}
\affil[3]{Tp-Link International Shenzhen Co.,Ltd., Shenzhen, China}
\begin{document}

\maketitle
\thispagestyle{empty}
\pagestyle{empty}

\begin{abstract}
Shadows in scanned documents pose significant challenges for document analysis and recognition tasks due to their negative impact on visual quality and readability. Current shadow removal techniques, including traditional methods and deep learning approaches, face limitations in handling varying shadow intensities and preserving document details. To address these issues, we propose DocDeshadower, a novel multi-frequency Transformer-based model built upon the Laplacian Pyramid. By decomposing the shadow image into multiple frequency bands and employing two critical modules—the Attention-Aggregation Network for low-frequency shadow removal and the Gated Multi-scale Fusion Transformer for global refinement—DocDeshadower effectively removes shadows at different scales while preserving document content. Extensive experiments demonstrate DocDeshadower's superior performance compared to state-of-the-art methods, highlighting its potential to significantly improve document shadow removal techniques. The code is available at \href{https://github.com/leiyingtie/DocDeshadower}{https://github.com/leiyingtie/DocDeshadower}.
\end{abstract}

\section{INTRODUCTION}
The proliferation of smartphone-based document digitization has revolutionized the way we capture and process documents. However, the presence of shadows and uneven illumination in smartphone-captured document images can significantly impair legibility and hinder the performance of various computer vision tasks, particularly segmentation~\cite{chen17,chen18,chen19} and optical character recognition~\cite{mori1999optical}.

  \begin{figure}[!ht]
    

    \begin{minipage}[b]{1.0\linewidth}
        \begin{minipage}[b]{.32\linewidth}
            \centering
            \centerline{\includegraphics[width=\linewidth]{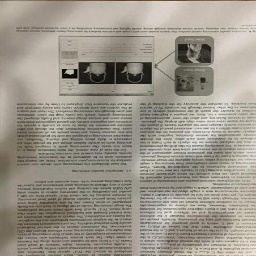}}
            \centerline{(a)Input}\medskip
        \end{minipage}
        \hfill
        \begin{minipage}[b]{0.32\linewidth}
            \centering
            \centerline{\includegraphics[width=\linewidth]{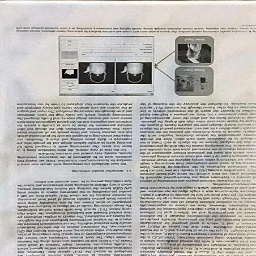}}
            \centerline{(b)Kligler \etal}\medskip
        \end{minipage}
        \hfill
        \begin{minipage}[b]{0.32\linewidth}
            \centering
            \centerline{\includegraphics[width=\linewidth]{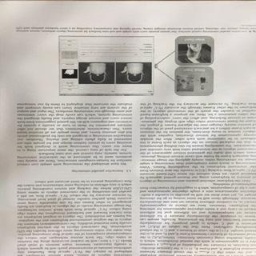}}
            \centerline{(c)DHAN}\medskip
        \end{minipage}
    \end{minipage}

    \begin{minipage}[b]{1.0\linewidth}
        \begin{minipage}[b]{0.32\linewidth}
            \centering
            \centerline{\includegraphics[width=\linewidth]{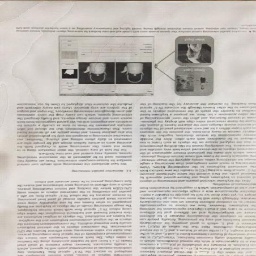}}
            \centerline{(d)SG-ShadowNet}\medskip
        \end{minipage}
        \hfill
                \begin{minipage}[b]{0.32\linewidth}
            \centering
            \centerline{\includegraphics[width=\linewidth]{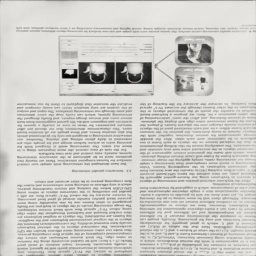}}
            \centerline{(e)Ours}\medskip
        \end{minipage}
        \hfill
        \begin{minipage}[b]{0.32\linewidth}
            \centering
            \centerline{\includegraphics[width=\linewidth]{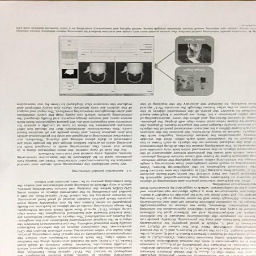}}
            \centerline{(f)Target}\medskip
        \end{minipage}
    \end{minipage}
    \caption{
    The figure illustrates the comparison between our method and several state-of-the-art methods in document shadows removal.
    }
  \label{fig:teaser}
  \end{figure}

Traditional document shadow removal techniques rely heavily on heuristics to analyze image features~\cite{bako2017removing,shah2018iterative,kligler2018document,jung2019water,wang2019effective,wang2020shadow,liu2023shadow}. These methods typically focus on detecting intensity changes or examining illumination in digitized documents. However, these heuristic-based approaches have limitations, as they may perform well for certain document images but fail to generalize effectively to others, as illustrated in Figure~\ref{fig:teaser} (b).

In recent years, the advent of Convolutional Neural Networks (CNN) and Generative Adversarial Networks (GANs)\cite{Goodfellow2014GenerativeAN,chen10,chen11,liu1} has revolutionized the field of image processing, leading to the development of deep learning approaches for shadow removal in both natural and document images\cite{wang2018stacked,hu2019mask,lin2020bedsr,cun2020towards,le2020shadow,fu2021auto,liu2021shadow,jin2021dc,wan2022style,guo2023shadowformer}. These techniques leverage paired shadow images, shadow-free images, and shadow masks to detect and eliminate shadows through end-to-end training. However, the requirement for a substantial volume of paired document images poses a challenge. Although researchers have proposed real datasets consisting of shadow-free and shadowed image pairs, the high cost associated with constructing large datasets has resulted in the availability of only a limited number of smaller real datasets~\cite{kligler2018document,jung2019water}. Consequently, the trained networks may struggle to generalize effectively to diverse document types, as depicted in Figure~\ref{fig:teaser} (c) and (d).

To address these challenges, we propose a multi-scale method that takes into account the distinct manifestations of shadows across different frequency domains. We propose the Attention-Aggregation Network to handle shadow color information at the pixel level and the Multi-scale Fusion Transformer to refine shadow edge features on a global scale, leveraging the large perceptual field of the Vision Transformer~\cite{dosovitskiy2020vit,chen1,chen6,chen16}. This strategy enables effective handling of shadow characteristics across various frequencies and scales. We employ the Laplacian Pyramid~\cite{burt1987laplacian,chen2,chen3,chen4,chen5} for frequency decomposition, allowing the division of the image into multiple frequency bands and lossless reconstruction~\cite{chen7,chen8}. To mitigate the issue of data scarcity, we utilize various data augmentation techniques.

Our main contributions are as follows:
\begin{itemize}
\item We propose DocDeshadower, a novel Transformer-based network that addresses the multi-scale nature of shadows by employing the Laplacian pyramid to handle color distortion in low-frequency domain and edge features in high-frequency domain.
\item We propose the Attention-Aggregation Network for mitigating color distortion in the low-frequency component and the Gated Multi-scale Fusion Transformer for refining high-frequency edge features, enhancing the removal of shadows across various frequencies.
\item Through comprehensive benchmark experiments, we demonstrate that our model significantly surpasses the performance of state-of-the-art techniques.
\end{itemize}

\section{RELATED WORK}
Shadow removal, crucial for enhancing image quality and aiding computer vision tasks, has been extensively studied for both natural images and documents.
\subsection{Natural Image Shadow Removal}
Methods for removing shadows from natural images range from physics-based models to deep learning approaches. Early works like DeShadowNet~\cite{qu2017deshadownet} and ST-CGAN~\cite{wang2018stacked} employed convolutional neural networks to learn shadow characteristics and generate shadow-free images. More recent approaches leverage attention mechanisms, multi-scale analysis, and style transfer techniques for improved performance. For instance, DHAN~\cite{cun2020towards} aggregates multi-context features, while SG-ShadowNet~\cite{wan2022style} addresses inconsistencies between shadow and non-shadow regions using a style-guided approach.
\subsection{Document Shadow Removal}
Document shadow removal presents unique challenges due to the presence of text and the need to preserve fine details. Traditional methods often rely on statistical analysis and intrinsic image decomposition~\cite{bako2017removing, kligler2018document, shah2018iterative, jung2019water, wang2019effective, wang2020shadow}. BEDSR-Net~\cite{lin2020bedsr} marked the first deep learning approach specifically designed for document shadow removal. More recently, techniques incorporating text elimination and adaptive contrast enhancement have been proposed~\cite{liu2023shadow}.
\subsection{Foundational Techniques for DocDeshadower}
Our proposed DocDeshadower model builds upon three key foundational techniques:
\begin{itemize}
	\item Laplacian Pyramid~\cite{burt1987laplacian}: This multi-scale decomposition technique allows us to analyze and process document shadows at different levels of detail, effectively addressing their inherent multi-scale nature.
	\item Transformer Architectures~\cite{dosovitskiy2020vit,chen13}: By leveraging the self-attention mechanism of transformers, DocDeshadower captures long-range dependencies within the image, enabling a holistic understanding of the document context.
	\item Attention Mechanisms: Attention mechanisms~\cite{chen9,chen12,chen14} allow DocDeshadower to focus on the most relevant image regions, effectively distinguishing between shadowed and non-shadowed areas for precise shadow removal.
\end{itemize}
These techniques, combined in our novel architecture, enable DocDeshadower to effectively address the challenges of document shadow removal while preserving text integrity and image details.

\section{METHODOLOGY}
\begin{figure}[ht]
    \begin{minipage}[b]{1.0\linewidth}
        \centering
            \centerline{\includegraphics[width=\linewidth]{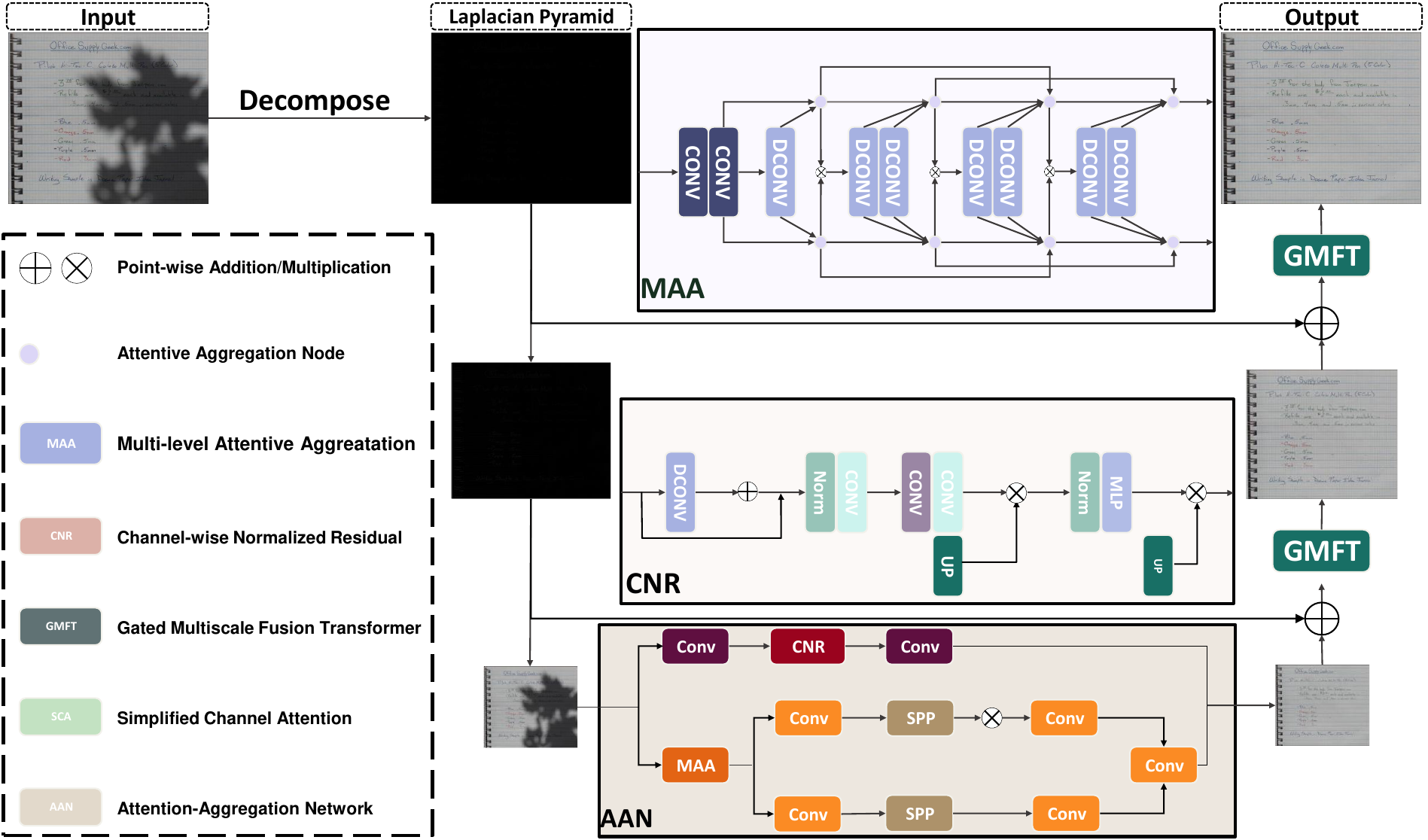}}
    \end{minipage}
    \caption{
    The proposed DocDeshadower architecture.
    }
    \label{fig:model}
\end{figure}

\subsection{Overview}
Shadows possess two types of features: color information in low-frequency and textural features in high-frequency components. Decomposing an image into various scales of low and high-frequency information enables the extraction of multi-scale details, thereby efficiently eliminating shadows. We disassemble shadow images into low and high-frequency components using the Laplacian Pyramid, which facilitates multi-scale processing and preserves detail and textural information more effectively. After decomposing the input image, we acquire multiple sections. The Gated Multi-scale Fusion Transformer (GMFT) handles the high-frequency components, while the Attention-Aggregation Network (AAN) processes the low-frequency component.

\subsection{Attention-Aggregation Network}
Inspired by~\cite{cun2020towards}, we propose the Attention-Aggregation Network to improve the low-frequency components of the image. The network consists of two branches, as shown in Figure~\ref{fig:model}.

The first branch encompasses the Channel-wise Normalized Residual (CNR) block, which effectively identifies shadows in the image, offering a crucial component in differentiating between foreground and background elements in shadow removal tasks. The CNR block intensifies the focus on developing features across channels. The output of the CNR block can be formulated as:
\begin{equation}
CNR(x) = \frac{x - \mu_c}{\sqrt{\sigma_c^2 + \epsilon}} \times \gamma + \beta,
\end{equation}
where $x$ is the input feature map, $\mu_c$ and $\sigma_c^2$ are the mean and variance of the feature map along the channel dimension, $\epsilon$ is a small constant for numerical stability, and $\gamma$ and $\beta$ are learnable parameters.

The second branch includes the Multi-stage Attentive Aggregation (MAA) block, which executes cascaded convolution and fusion. The Attentive Aggregation Node independently fuses features and attention. The Multilayer Perceptron (MLP) serves as a fundamental model for performing nonlinear transformations on features. The Spatial Pyramid Pooling (SPP) further enhances the model by pooling features at various stages, augmenting the breadth and depth of perception. The output of the MAA block can be expressed as:
\begin{equation}
MAA(x) = Concat(MLP(x), SPP(x)) \otimes Attention(x),
\end{equation}
where $\otimes$ denotes element-wise multiplication, and $Attention(x)$ represents the attention weights computed from the input feature map $x$.

\begin{figure}[ht]
    \begin{minipage}[b]{1.0\linewidth}
        \centering
            \centerline{\includegraphics[width=\linewidth]{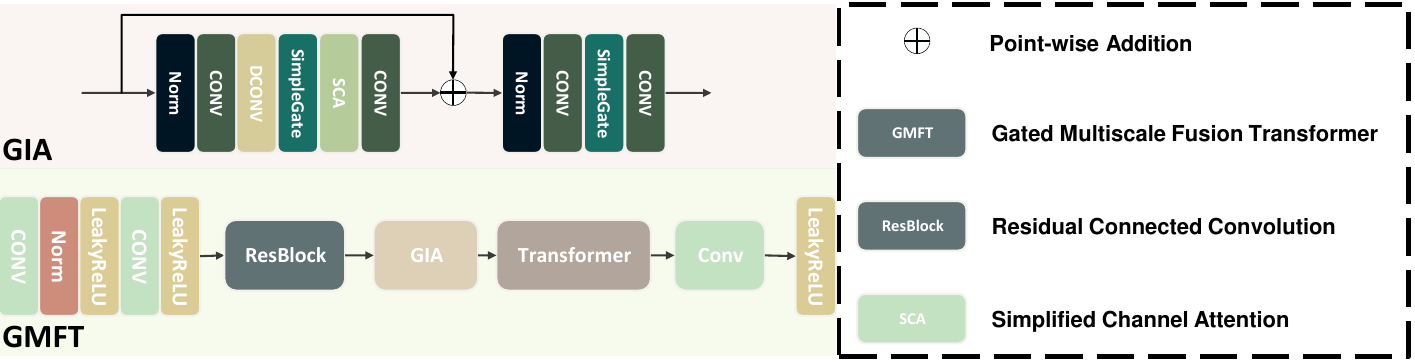}}
    \end{minipage}
    \caption{The proposed Gated Multi-scale Fusion Transformer architecture.
    }
    \label{fig:GMFT}
\end{figure}

\subsection{Gated Multi-scale Fusion Transformer}
We propose a novel Gated Multi-scale Fusion Transformer for processing the high-frequency component of the Laplacian Pyramid decomposition, as shown in Figure~\ref{fig:GMFT}. The input first undergoes feature enhancement via a Residual Connected Convolution (ResBlock) before being inputted into a streamlined Gated Identity Attention (GIA) Block. The ResidualBlock boosts both the accuracy and training speed of the model, while the GIA Block adjusts features in the color channel, enhances model performance, and decreases model parameters and computational complexity. The output of the GIA block can be formulated as:
\begin{equation}
GIA(x) = x \odot \sigma(W_g x) + b_g,
\end{equation}
where $\odot$ denotes element-wise multiplication, $\sigma$ is the sigmoid activation function, and $W_g$ and $b_g$ are learnable parameters.

To further reduce computational complexity, we apply SimpleGate~\cite{testa2022aggregate}, introducing nonlinearity. Additionally, we employ simple channel attention to augment channel interaction and the textural quality of the output.

Subsequently, the input features are fused and minimized via a Transformer Block, enabling the model to concentrate on significant features while discarding noise and superfluous information. Drawing inspiration from~\cite{Wang_Zhang_Shen_Luo_Stenger_Lu_2022}, we incorporate the Bi-directional Multi-head Transformer (BMT), which includes self-attention modules. This bidirectional attention mechanism facilitates the capture of essential information in the feature map. The self-attention mechanism can be expressed as:
\begin{equation}
Attention(Q, K, V) = softmax(\frac{QK^T}{\sqrt{d_k}})V,
\end{equation}
where $Q$, $K$, and $V$ are the query, key, and value matrices, respectively, and $d_k$ is the dimension of the key vectors.

Lastly, we employ the Dual Gated Feed-forward Network (DGFN) with gating mechanisms to enact non-linear transformations on the input feature maps, leading to improved global adjustment. The DGFN can be formulated as:
\begin{equation}
DGFN(x) = W_2(ReLU(W_1x + b_1) \odot \sigma(W_gx + b_g)) + b_2,
\end{equation}
where $W_1$, $W_2$, $W_g$, $b_1$, $b_2$, and $b_g$ are learnable parameters, $ReLU$ is the rectified linear unit activation function, and $\sigma$ is the sigmoid activation function.

\subsection{Objective Functions}
Our objective function incorporates the Mean Squared Error ($L_{MSE}$) and the Structural Similarity Index ($L_{SSIM}$) to formulate a comprehensive loss function, $L_{total}$, as delineated in Equation~\ref{eq:total}:
\begin{equation}
L_{total} = L_{MSE} + \lambda \times L_{SSIM},
\label{eq:total}
\end{equation}
where $\lambda$ is empirically set to 0.2.

The Mean Squared Error, $L_{MSE}$, is defined as:
\begin{equation}
L_{MSE}=\sum_{i=1}^{n}(x_i-y_i)^2,
\label{eq:mse}
\end{equation}
where $x_i$ and $y_i$ represent the pixel values of the predicted and ground truth images, respectively.

The Structural Similarity Index, $L_{SSIM}$, is calculated as:
\begin{equation}
L_{SSIM}=\frac{\left(2 \mu_{\mathrm{x}} \mu_{\mathrm{y}}+C_{1}\right)\left(2 \sigma_{\mathrm{xy}}+C_{2}\right)}{\left(\mu_{\mathrm{x}}^{2}+\mu_{\mathrm{y}}^{2}+C_{1}\right)\left(\sigma_{\mathrm{x}}^{2}+\sigma_{\mathrm{y}}^{2}+C_{2}\right)},
\label{eq:ssim}
\end{equation}
where $\mu_{\mathrm{x}}$ and $\mu_{\mathrm{y}}$ denote the means of images X and Y, $\sigma_{\mathrm{xy}}$ represents the covariance between images X and Y, and $\sigma_{\mathrm{x}}$ and $\sigma_{\mathrm{y}}$ denote the standard deviations of images X and Y, respectively. The constants $C_1$ and $C_2$ are set to $(0.01 \times 255)^{2}$ and $(0.03 \times 255)^{2}$, respectively.

\begin{figure*}[ht]
    \begin{minipage}[b]{1.0\linewidth}
        \begin{minipage}[b]{0.137\linewidth}
            \centering
            \centerline{\includegraphics[width=\linewidth]{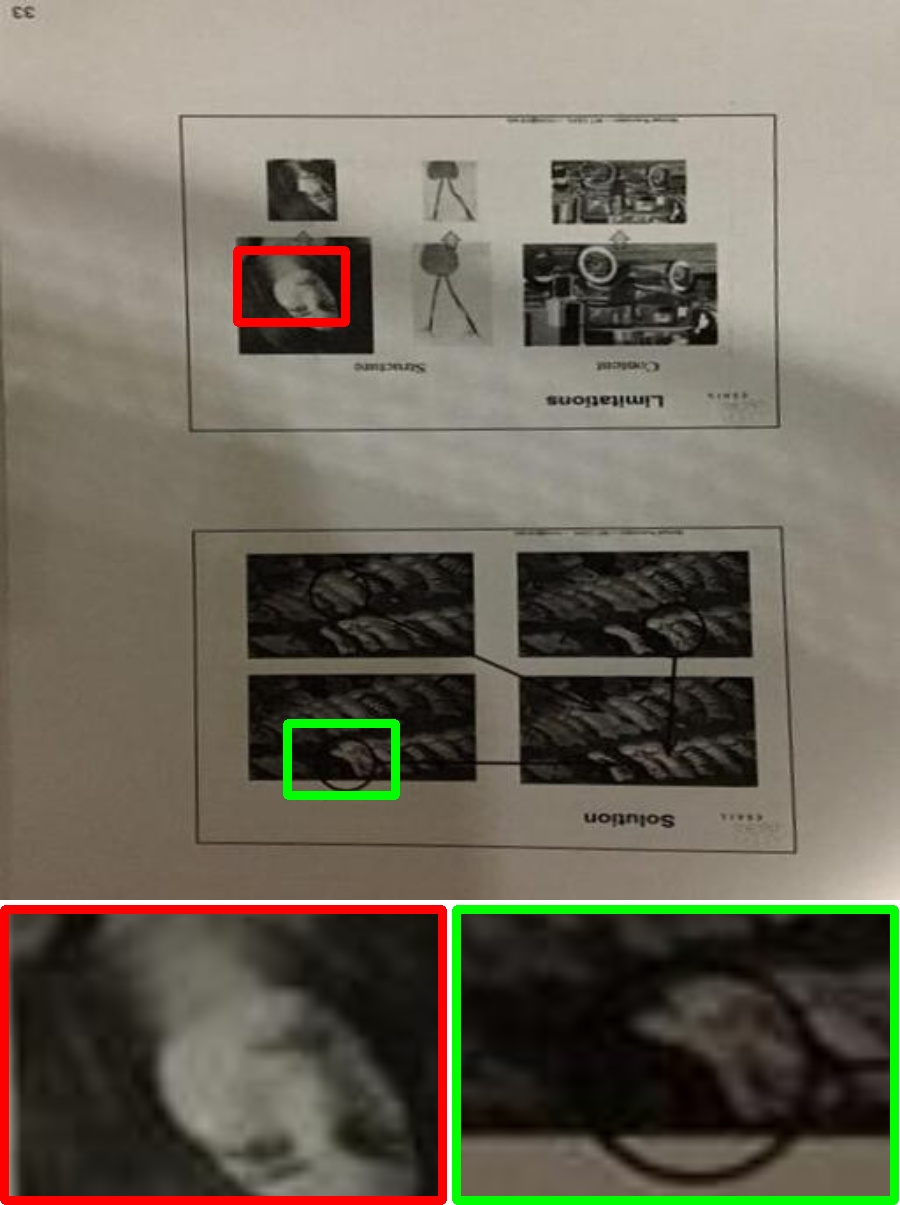}}
        \end{minipage}   
        \hfill
        \begin{minipage}[b]{0.137\linewidth}
            \centering
            \centerline{\includegraphics[width=\linewidth]{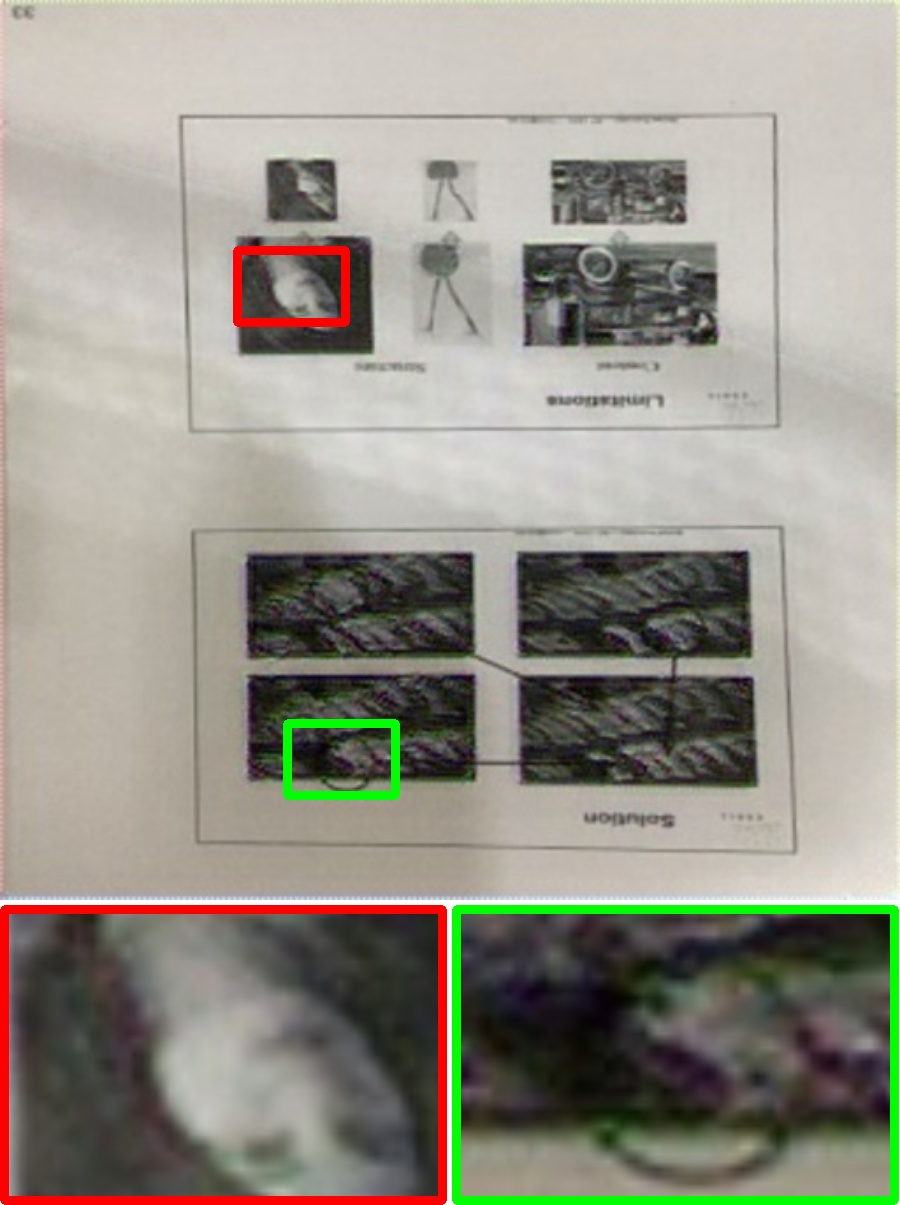}}
        \end{minipage}
        \hfill
        \begin{minipage}[b]{0.137\linewidth}
            \centering
            \centerline{\includegraphics[width=\linewidth]{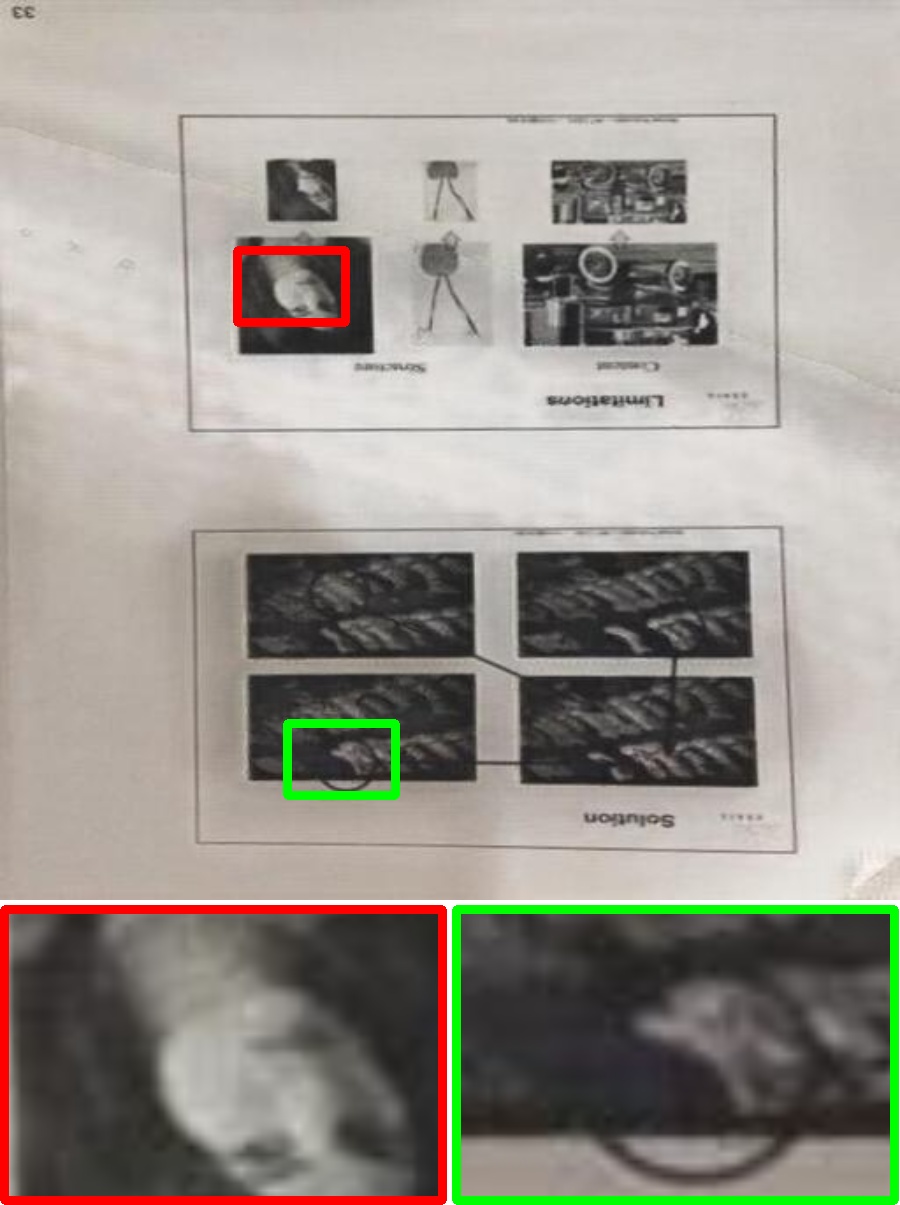}}
        \end{minipage}
        \hfill
        \begin{minipage}[b]{0.137\linewidth}
            \centering
            \centerline{\includegraphics[width=\linewidth]{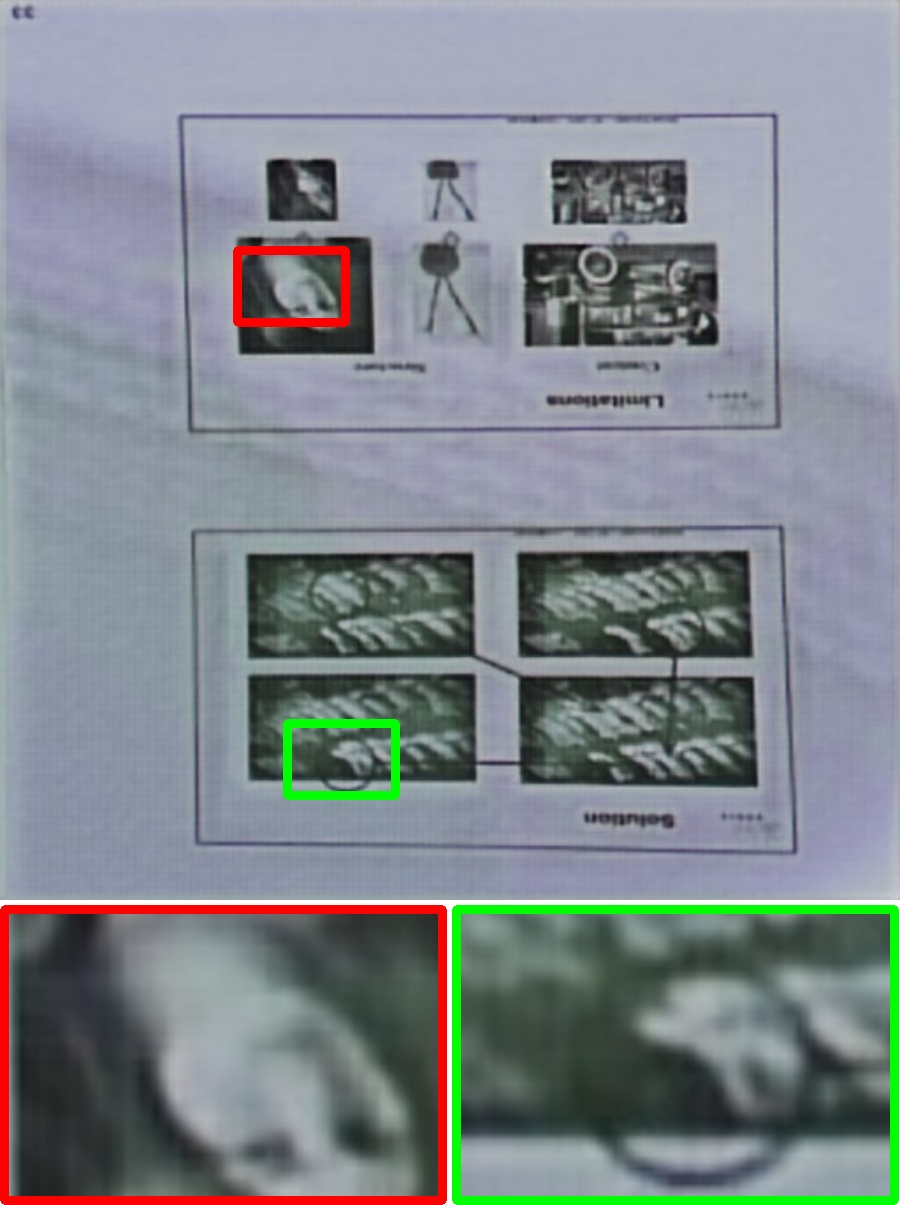}}
        \end{minipage}
        \hfill   
        \begin{minipage}[b]{0.137\linewidth}
            \centering
            \centerline{\includegraphics[width=\linewidth]{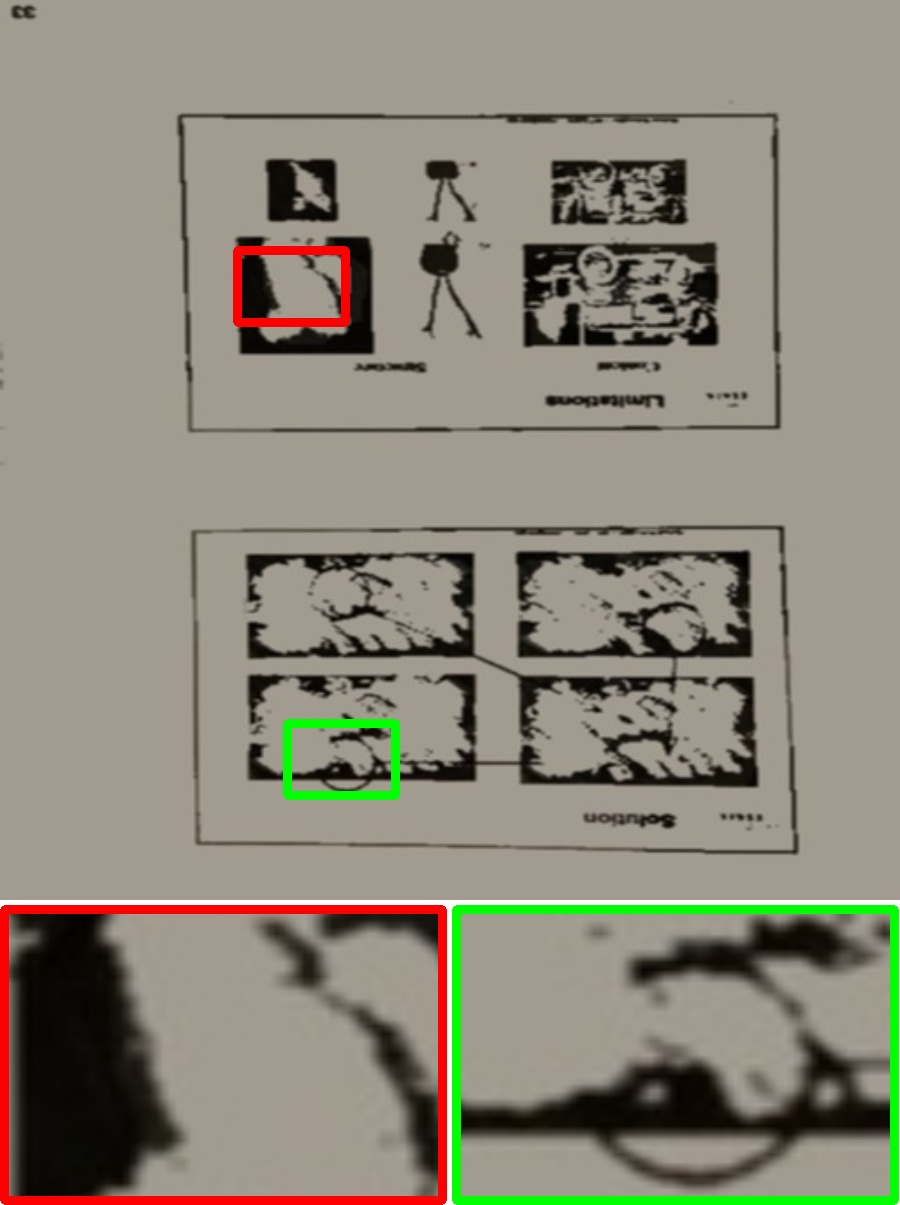}}
        \end{minipage}  
        \hfill
        \begin{minipage}[b]{0.137\linewidth}
            \centering
            \centerline{\includegraphics[width=\linewidth]{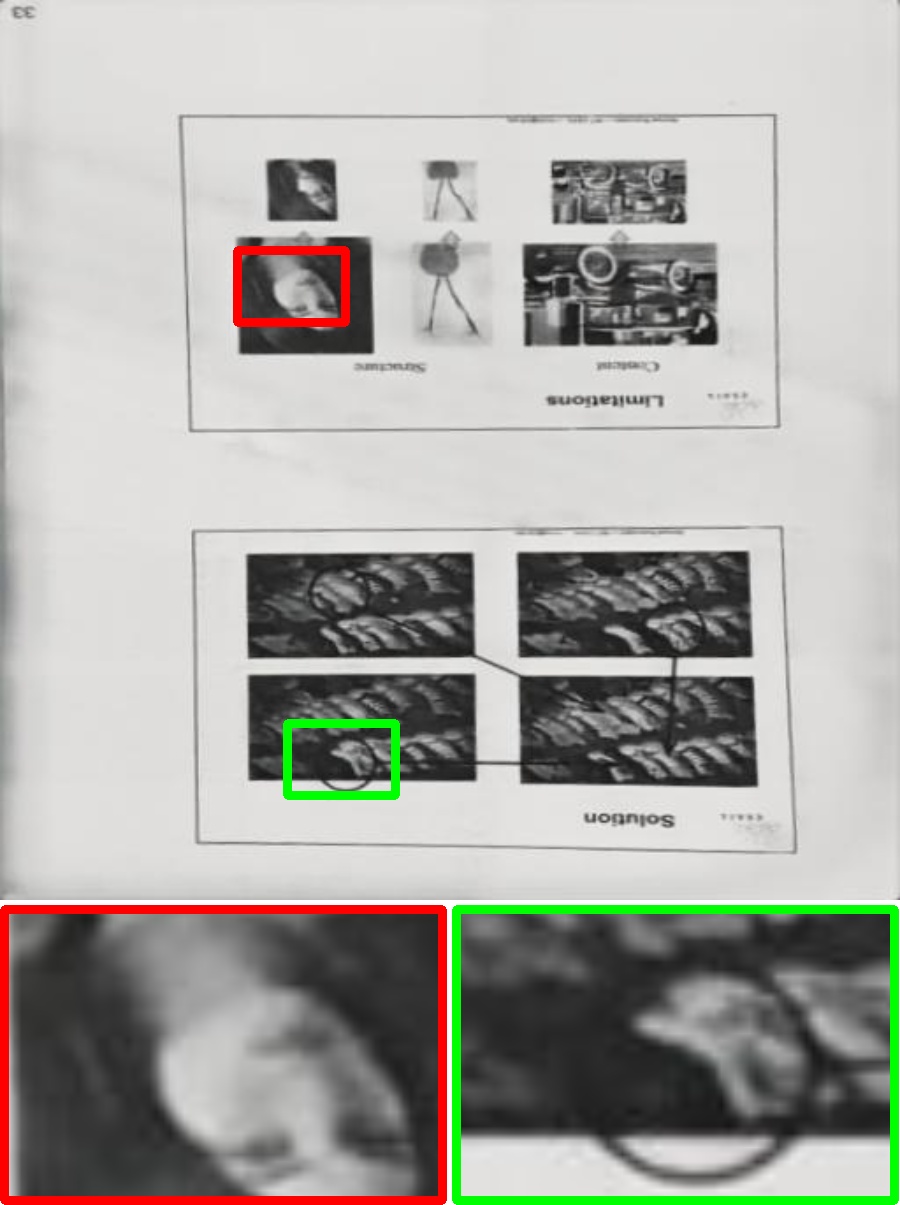}}
        \end{minipage}
        \hfill
        \begin{minipage}[b]{0.137\linewidth}
            \centering
            \centerline{\includegraphics[width=\linewidth]{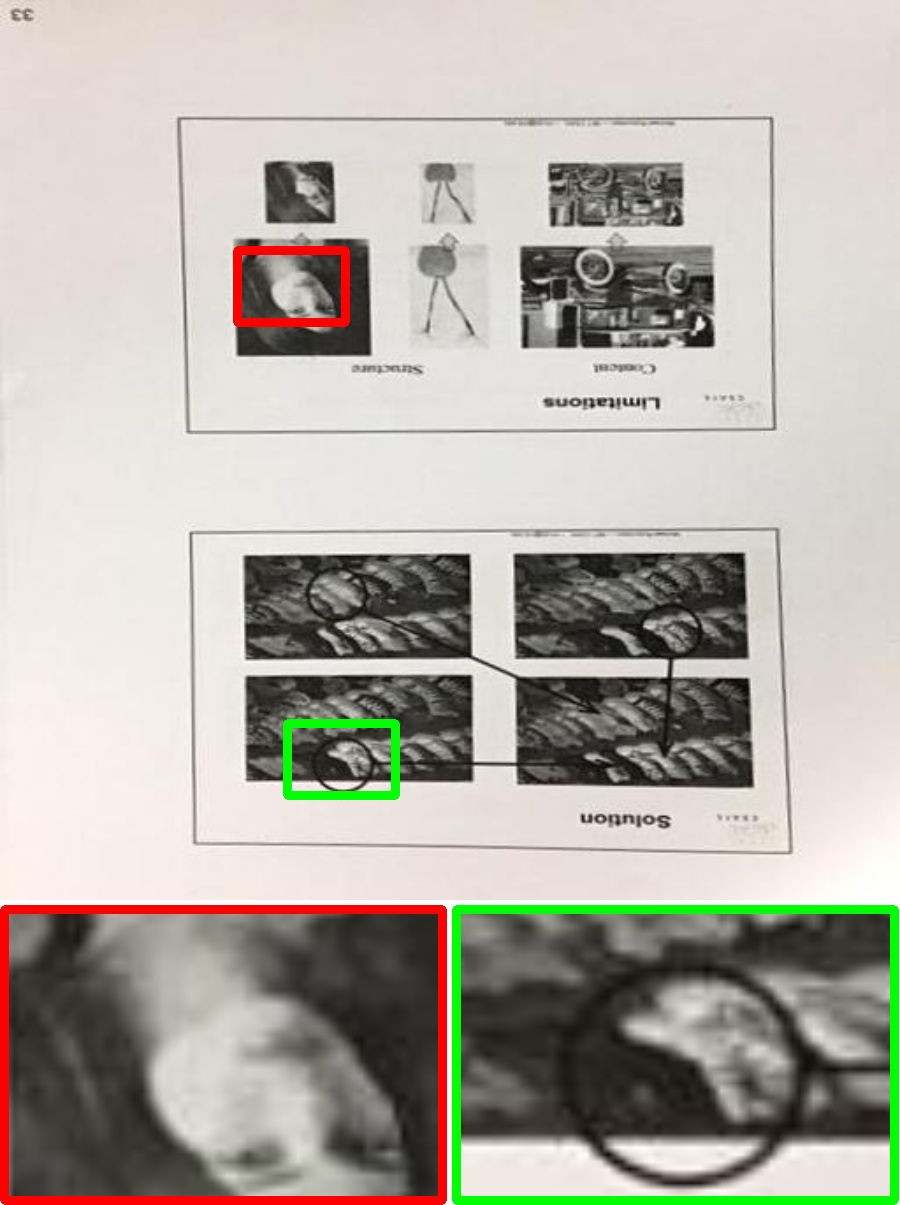}}
        \end{minipage}
    \end{minipage}
    \begin{minipage}[b]{1.0\linewidth}
        \begin{minipage}[b]{0.137\linewidth}
            \centering
            \centerline{\includegraphics[width=\linewidth]{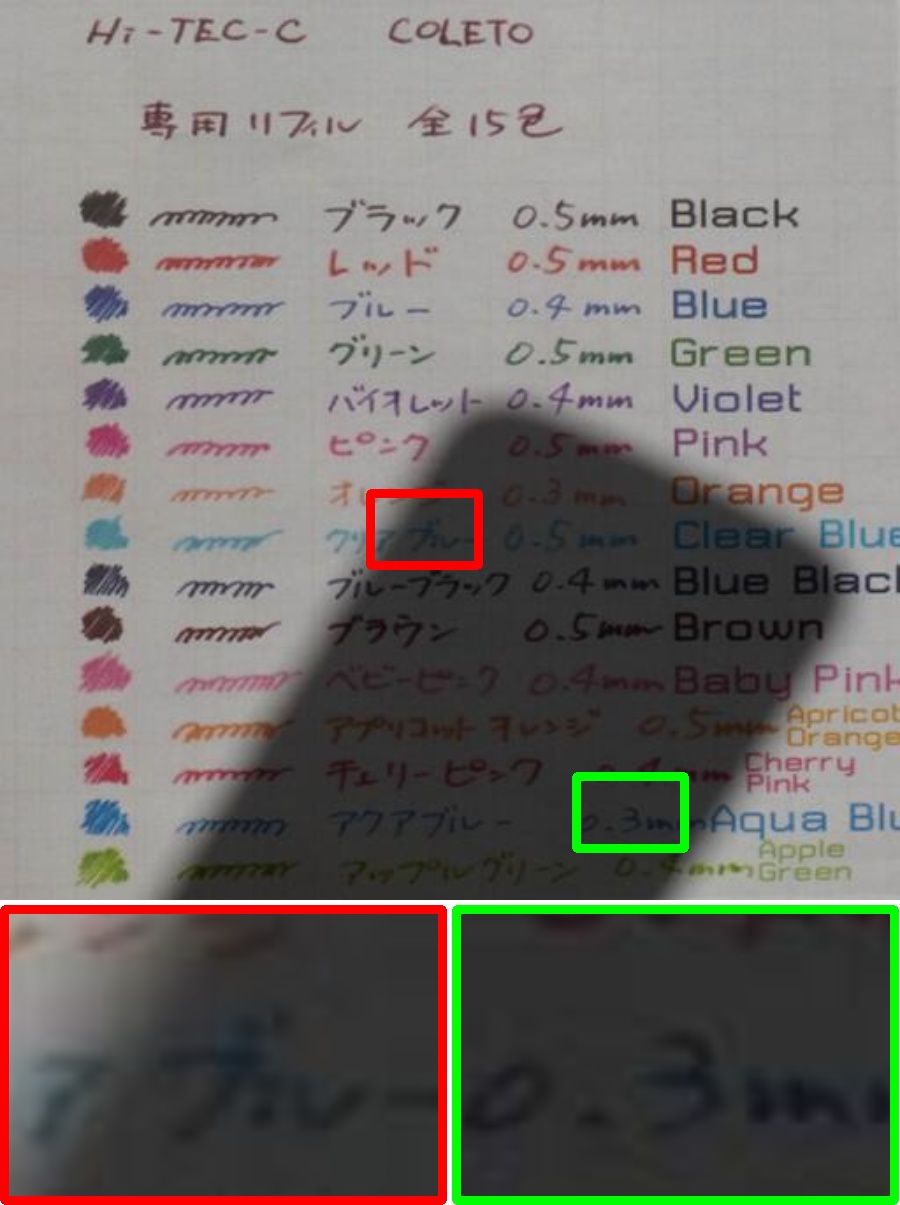}}
            \centerline{(a)Input}\medskip
        \end{minipage}   
        \hfill
        \begin{minipage}[b]{0.137\linewidth}
            \centering
            \centerline{\includegraphics[width=\linewidth]{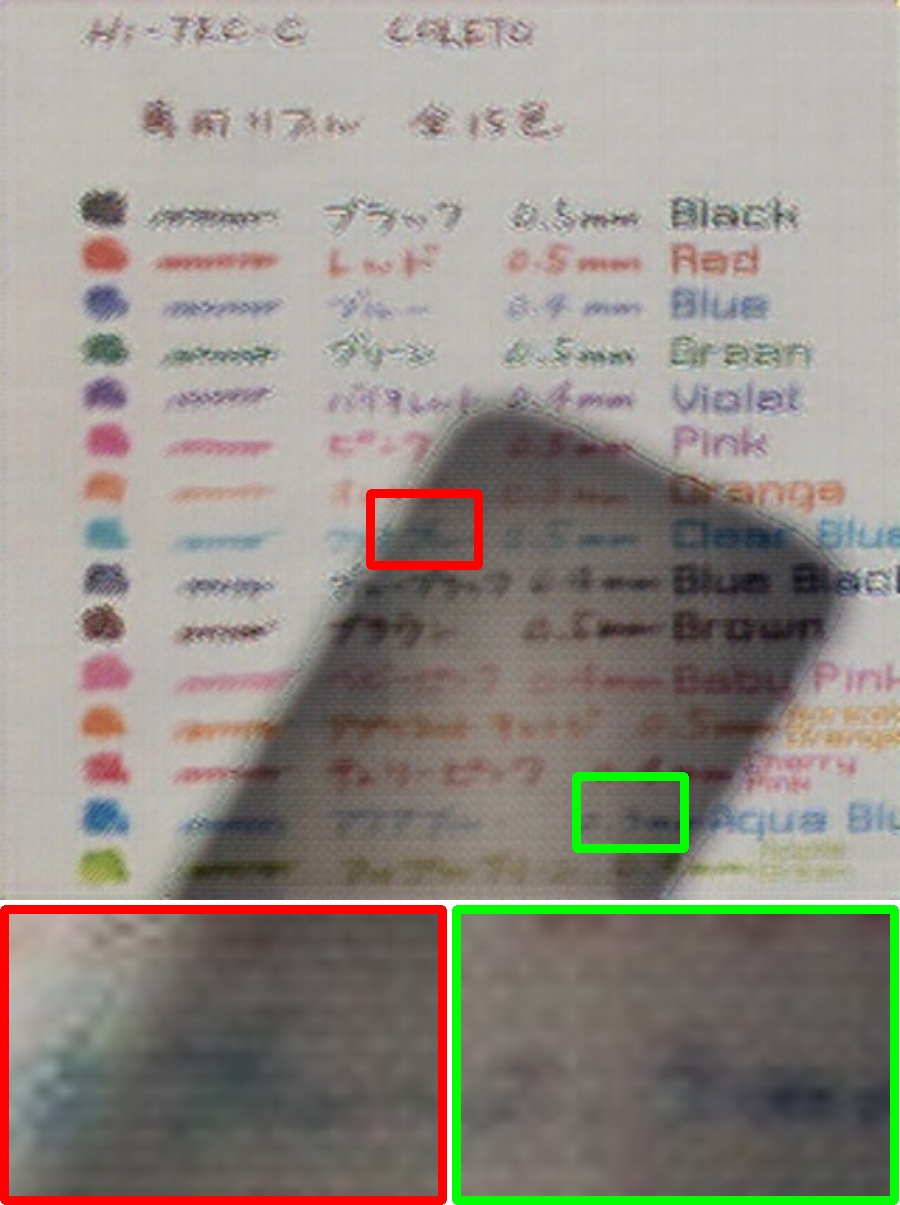}}
            \centerline{(b)ShadowFormer}\medskip
        \end{minipage}
        \hfill
        \begin{minipage}[b]{0.137\linewidth}
            \centering
            \centerline{\includegraphics[width=\linewidth]{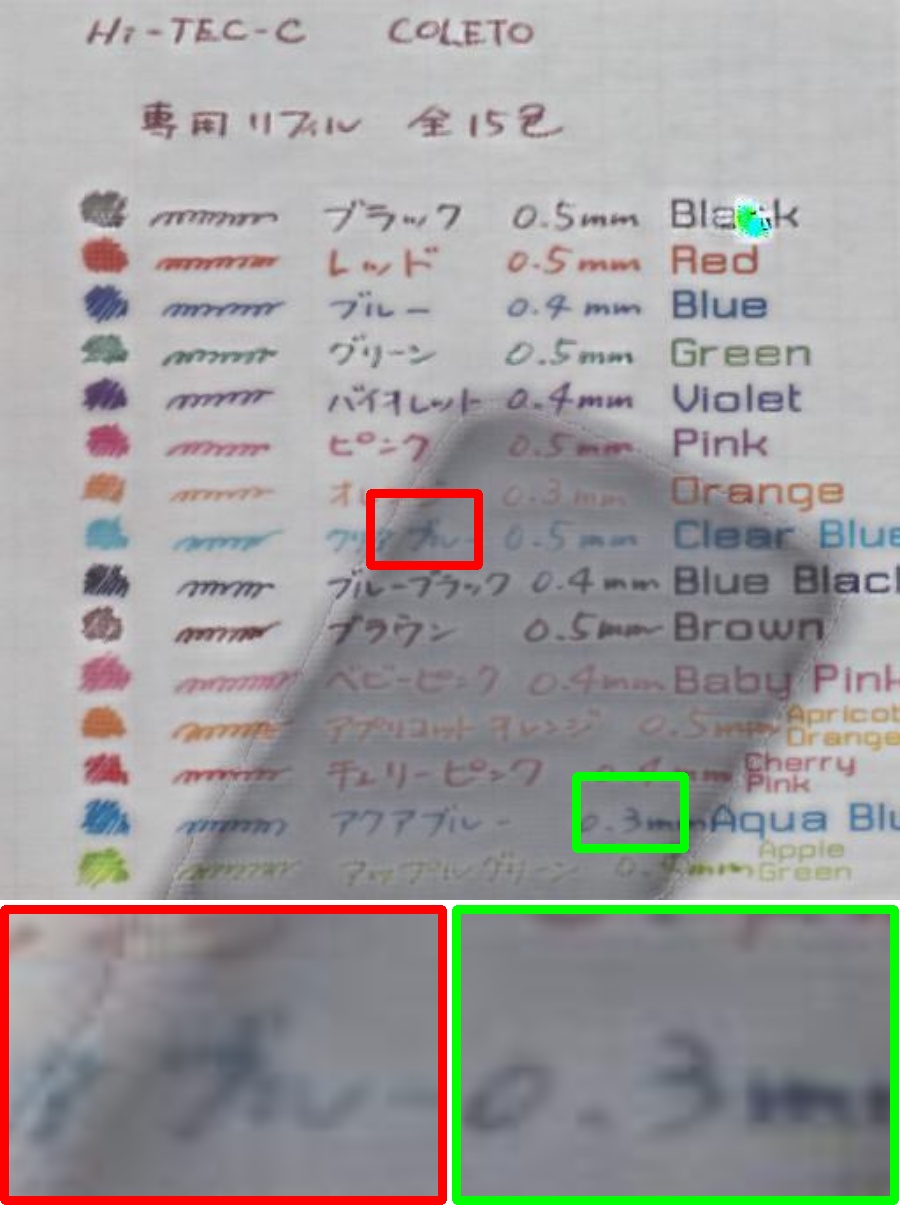}}
            \centerline{(c)SG-ShadowNet}\medskip
        \end{minipage}
        \hfill
        \begin{minipage}[b]{0.137\linewidth}
            \centering
            \centerline{\includegraphics[width=\linewidth]{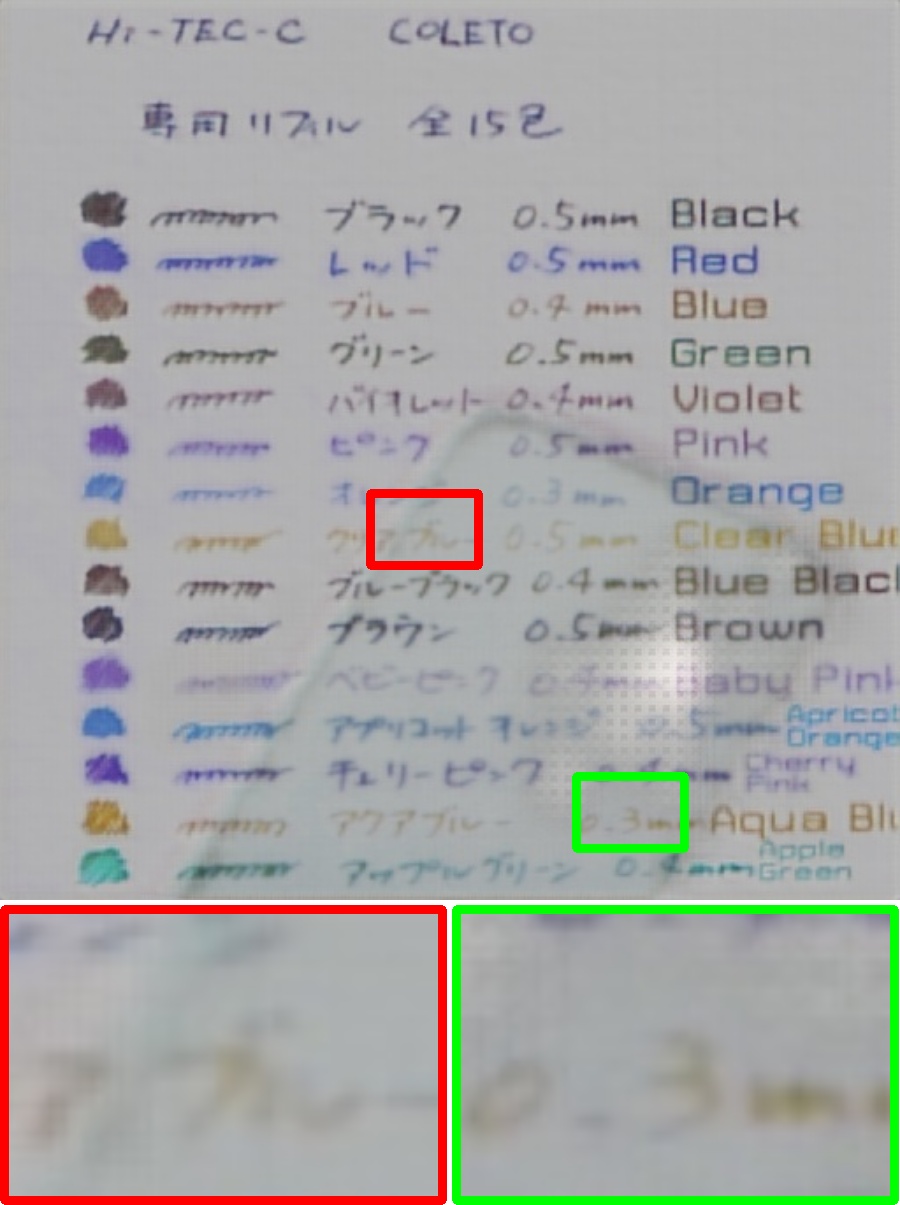}}
            \centerline{(d)BEDSR-Net}\medskip
        \end{minipage}
        \hfill   
        \begin{minipage}[b]{0.137\linewidth}
            \centering
            \centerline{\includegraphics[width=\linewidth]{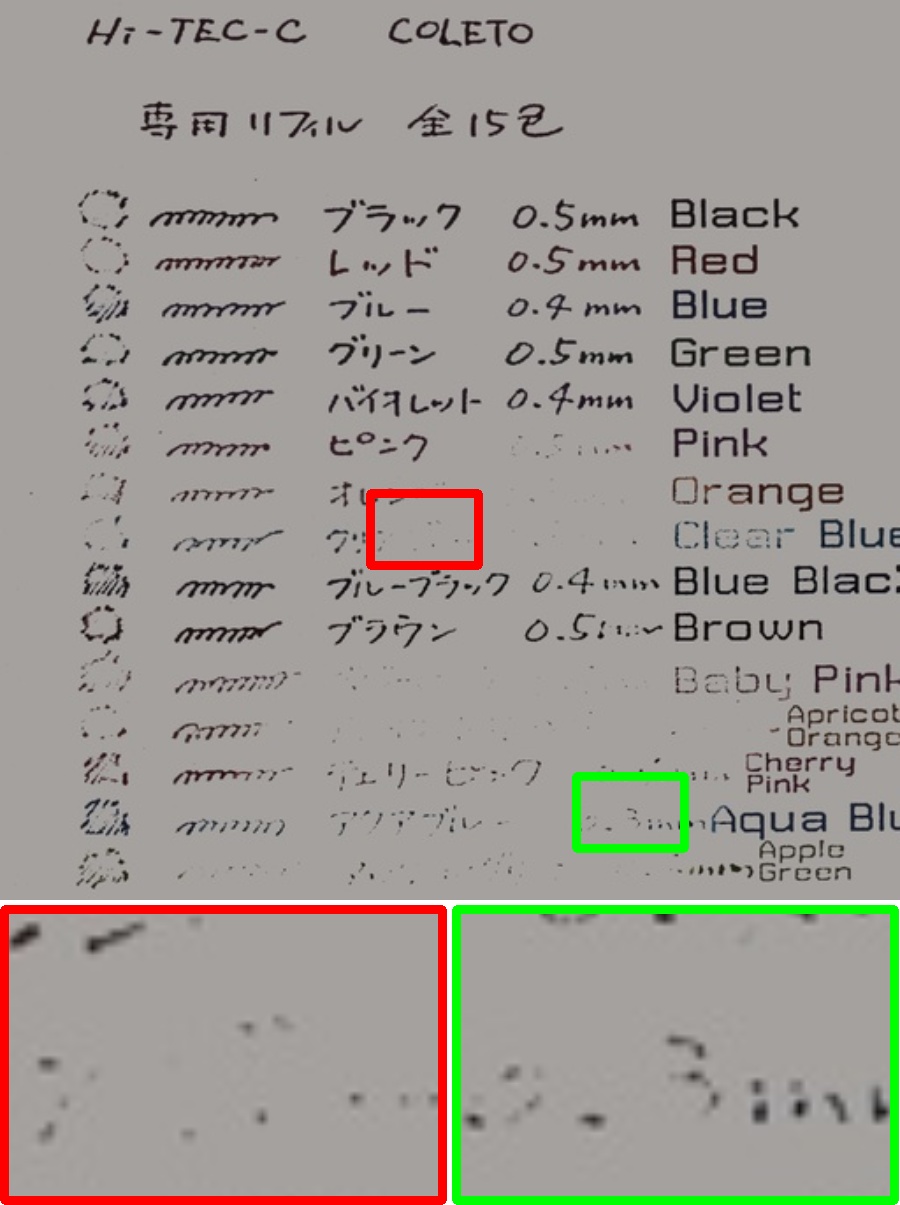}}
            \centerline{(e)Liu \etal}\medskip
        \end{minipage}  
        \hfill
        \begin{minipage}[b]{0.137\linewidth}
            \centering
            \centerline{\includegraphics[width=\linewidth]{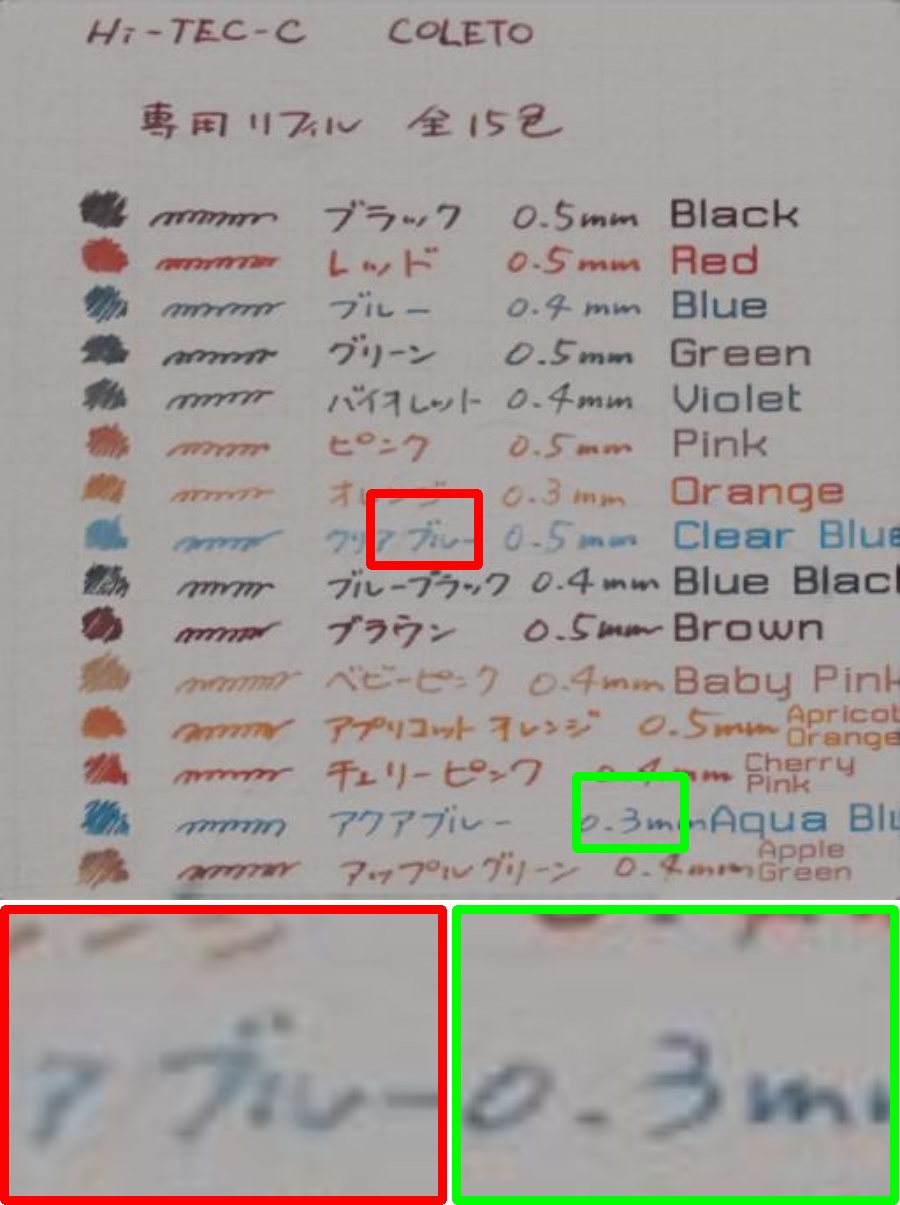}}
            \centerline{(f)Ours}\medskip
        \end{minipage}
        \hfill
        \begin{minipage}[b]{0.137\linewidth}
            \centering
            \centerline{\includegraphics[width=\linewidth]{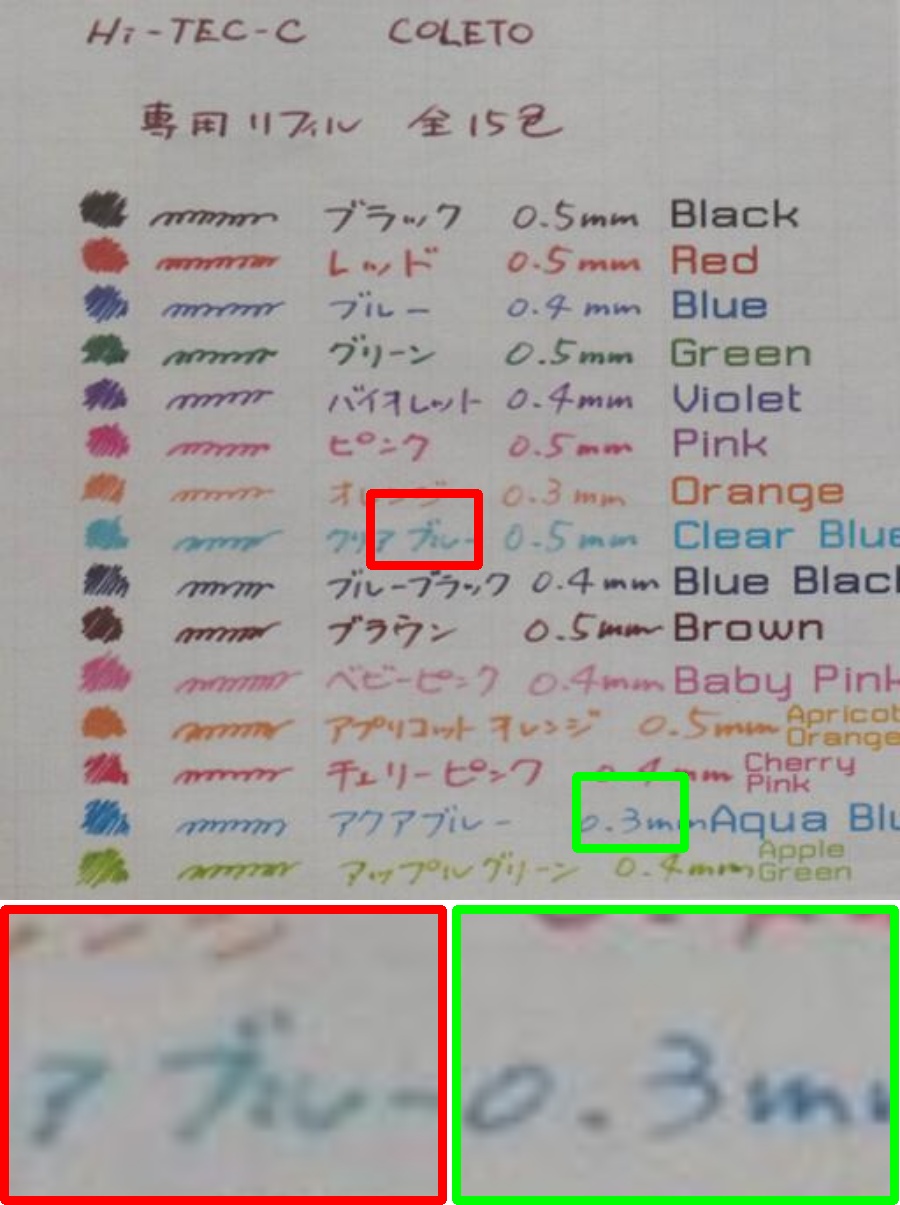}}
            \centerline{(g)Target}\medskip
        \end{minipage}
    \end{minipage}
    \caption{
    Visual results of different document shadow removal methodologies.}
    \label{fig:ref}
\end{figure*}

\section{EXPERIMENTS}

\subsection{Experimental Setup}

\begin{table}[ht]
\caption{Details of Jung and Kligler dataset.}
\centering
\begin{tabular}{lccc}
\hline
Dataset & \# of training & \# of testing & Resolution \\ \hline
Jung~\cite{jung2019water}    & 60             & 27            & $512\times512$  \\
Kligler~\cite{kligler2018document} & 272            & 28            & $512\times 512$  \\ \hline
\end{tabular}
\label{tab:ds}
\end{table}
\subsubsection{Datasets and Preprocessing}
We benchmark the performance of each method on two publicly accessible datasets. The datasets employed are shown in Table~\ref{tab:ds}.

\subsubsection{Evaluation Metrics}
To evaluate the performance of DocDeshadower, we use three standard metrics:
\begin{itemize}
	\item Peak Signal-to-Noise Ratio (PSNR): Measures the ratio between the maximum possible signal power and the power of distorting noise. Higher PSNR values indicate better quality of the reconstructed image.
	\item Structural Similarity Index (SSIM): Assesses the perceived quality of digital images and videos. SSIM values closer to 1 indicate greater structural similarity to the ground truth.
	\item Root Mean Square Error (RMSE): Measures the standard deviation of residuals (prediction errors). Lower RMSE values indicate better fit to the ground truth.
\end{itemize}

\subsubsection{Implementation Details}
Our model is implemented using PyTorch and trained on an NVIDIA RTX Titan GPU using the default parameters of the standard Adam optimizer. The batch size and learning rate are set to 1 and $1e-4$, respectively. We augment the training dataset through random cropping, resizing, flipping, mixup, as well as brightness and saturation adjustments.

\begin{table}[ht]
\centering
\caption{Quantitative results of comparisons with the state-of-the-art methods on Jung and Kligler datasets.}
\adjustbox{width=\columnwidth}{
\begin{tabular}{l|ccc|ccc}
\hline
\multirow{2}{*}{Method} &       & Jung &       &       & Kligler &       \\ \cline{2-7} 
                        & PSNR$\uparrow$  & SSIM$\uparrow$ & RMSE$\downarrow$  & PSNR$\uparrow$  & SSIM$\uparrow$    & RMSE$\downarrow$  \\ \hline
Input                   & 13.01 & 0.82 & 60.85 & 13.26 & 0.8     & 56.73 \\
Shah \etal~\cite{shah2018iterative}                    & 14.68 & 0.8 & 47.98 & 8.36 & 0.7 & 97.88 \\
Kligler \etal~\cite{kligler2018document}                 & 18.56 & 0.81 & 31.18 & 18.15 & 0.81    & 32.4  \\
Jung \etal~\cite{jung2019water}                    & 22.39 & 0.87 & 19.94 & 14.34 & 0.83    & 49.73 \\
Wang \etal~\cite{wang2019effective}              & 9.11 & 0.71 & 90.97 & 15.36 & 0.72    & 48.1\\
Wang \etal~\cite{wang2020shadow}              & 11.17  & 0.78 & 73.25 & 15.73 & 0.82    & 44.05\\
Liu \etal~\cite{liu2023shadow}             & 15.31  & 0.81 & 47.67 & 20.61 & 0.78    & 24.29\\
DeShadowNet~\cite{qu2017deshadownet}            & 17.06 & 0.78 & 36.12 & 16.62 & 0.23    & 37.68 \\
ST-CGAN~\cite{wang2018stacked}                 & 14.05 & 0.33 & 52.02 & 12.3  & 0.44    & 62.23 \\
Mask-ShadowGAN~\cite{hu2019mask}          & 19.92 & 0.83 & 27.25 & 26.47 & 0.88    & 14.01 \\
BEDSR-Net~\cite{lin2020bedsr}               & 13.19 & 0.76 & 56.7  & 22.03 & 0.77    & 22.13 \\
DHAN~\cite{cun2020towards}                    & 20.48 & 0.82 & 26.28 & 25.58 & 0.84    & 15.89 \\
SP+M Net~\cite{le2020shadow}                & 23.18 & 0.86 & 20.69 & 25.4  & 0.88    & 14.95 \\
AEFNet~\cite{fu2021auto}                  & 19.19 & 0.83 & 29.96 & 27.1  & 0.91    & 14.52 \\
LG-ShadowNet~\cite{liu2021shadow}            & 19.68 & 0.83 & 28.59 & 20.04 & 0.85    & 26.16 \\
DC-ShadowNet~\cite{jin2021dc}            & 21.06 & 0.87 & 24.32 & 26.88 & 0.9     & 13.76 \\
SG-ShadowNet~\cite{wan2022style}            & 21.45 & 0.86 & 23.25 & 18.5  & 0.81    & 31.12 \\
Unfolding~\cite{zhu2022efficient}            & 21.81 & 0.86 & 23.01 & 26.89 & 0.88    & 12.53 \\
ShadowFormer~\cite{guo2023shadowformer}            & 20.13 & 0.82 & 27.85 & 16.69 & 0.65    & 37.87 \\
DMTN~\cite{dmtn}            & 22.3 & 0.85 & 20.47 & 26.92 & 0.88    & 13.96 \\
TBRNet~\cite{tbr}            & 21.71 & 0.85 & 22.03 & 26.3 & 0.88    & 14.28 \\
Ours                     & {\color[HTML]{FE0000} \textbf{23.52}} & {\color[HTML]{FE0000} \textbf{0.88}} & {\color[HTML]{FE0000} \textbf{17.56}} & {\color[HTML]{FE0000} \textbf{28.96}} & {\color[HTML]{FE0000} \textbf{0.93}} & {\color[HTML]{FE0000} \textbf{10.95}} \\ \hline
\end{tabular}
}
\label{t1:sota}
\end{table}
\subsection{Comparisons with State-of-the-Arts}
We compare the proposed network to various state-of-the-art methods, including traditional methods and deep learning-based methods. As some training codes are not publicly available, we re-implement those models based on their original papers.

\subsubsection{Quantitative Comparison}
The quantitative results in Table~\ref{t1:sota} demonstrate the superior performance of DocDeshadower across all metrics on both datasets. Notably, our method achieves significant improvements in PSNR and SSIM while reducing RMSE compared to both traditional and learning-based methods. The substantial improvement in PSNR (\eg, 3.83 dB increase over the next best method on the Jung dataset) indicates that DocDeshadower produces shadow-removed images that are closer to the ground truth in terms of pixel-level accuracy. This is crucial for maintaining the readability and clarity of document content. The high SSIM scores (0.87 and 0.81 for Jung and Kligler datasets, respectively) suggest that our method preserves the structural information of the document better than other approaches. This is particularly important for maintaining the integrity of text and graphical elements in documents. The lower RMSE values achieved by our method indicate reduced overall error in shadow removal, suggesting more consistent performance across different shadow conditions and document types.

These improvements can be attributed to the multi-scale approach of DocDeshadower, which allows it to handle both global color distortions and local textural changes caused by shadows. The attention mechanisms in both AAN and GMFT modules contribute to the model's ability to focus on relevant features for shadow removal while preserving document details.

\subsubsection{Qualitative Comparison}
The qualitative results in Figure~\ref{fig:ref} further support these quantitative findings. Other methods often leave residual shadow edges and remnants of the original shadow in substantial areas, along with a noticeable disparity in luminosity between the target and result images. In contrast, our method effectively eliminates shadow edges while preserving the luminosity of the target image.

\subsection{Theoretical Analysis of Model Performance}
The superior performance of DocDeshadower over baseline methods can be attributed to several key factors:
\begin{enumerate}
	\item Multi-scale approach: By leveraging the Laplacian Pyramid decomposition, our model addresses shadows at different frequency levels. This is crucial because document shadows manifest differently across frequency bands - color distortions in low frequencies and edge artifacts in high frequencies. Traditional methods often struggle to handle this multi-scale nature of shadows effectively.
	\item Specialized attention mechanisms: The Attention-Aggregation Network (AAN) for low-frequency components and the Gated Multi-scale Fusion Transformer (GMFT) for high-frequency components allow our model to focus on relevant features. The AAN effectively mitigates color distortions, while the GMFT refines edge features, enabling comprehensive shadow removal while preserving document details.
	\item Complementary modules: The combination of AAN and GMFT allows for a holistic approach to shadow removal. While the AAN handles global color changes in the low-frequency domain, the GMFT addresses local textural changes in the high-frequency domain. This complementary design ensures that shadows are removed effectively across different scales and frequency bands.
	\item Transformer architecture: The use of transformer-based modules, particularly in the GMFT, enables our model to capture long-range dependencies in the image. This is especially beneficial for document images where the shadow effect can span large areas and have complex patterns.
\end{enumerate}

These design choices allow DocDeshadower to outperform existing methods, particularly in preserving document details while effectively removing shadows across various scales and intensities.

\begin{table}[ht]
\centering
\caption{Ablation studies on Jung and Kligler datasets.}
\adjustbox{width=\columnwidth}{
\begin{tabular}{l|ccc|ccc}
\hline
                         &                                       & Jung                                 &                                       &                                       & Kligler                              &                                       \\ \cline{2-7} 
\multirow{-2}{*}{Method} & PSNR$\uparrow$                                  & SSIM$\uparrow$                                 & RMSE$\downarrow$                                  & PSNR$\uparrow$                                  & SSIM$\uparrow$                                 & RMSE$\downarrow$                                  \\ \hline
Ours w/o AAN           & 22.54 & 0.8 & 19.17 & 24.83 & 0.83 & 15.59
                              \\
Ours w/o GMFT          & 21.11 & 0.79 & 22.75 & 22.9 & 0.81 & 18.94  \\
Ours                     & {\color[HTML]{FE0000} \textbf{23.52}} & {\color[HTML]{FE0000} \textbf{0.88}} & {\color[HTML]{FE0000} \textbf{17.56}} & {\color[HTML]{FE0000} \textbf{28.96}} & {\color[HTML]{FE0000} \textbf{0.93}} & {\color[HTML]{FE0000} \textbf{10.95}} \\ \hline
\end{tabular}
}
\label{t2:ablation}
\end{table}
\subsection{Ablation Studies}
To evaluate the effectiveness of our proposed DocDeshadower method, we conduct ablation studies testing several components, including:

\begin{enumerate}
\item \textbf{Ours w/o AAN}: the proposed model without the Attention-Aggregation Network (AAN).
\item \textbf{Ours w/o GMFT}: the proposed model without the Gated Multi-scale Fusion Transformer (GMFT).
\item \textbf{Ours}: the full model architecture.
\end{enumerate}

Table~\ref{t2:ablation} presents the results of our ablation studies on the Jung and Kligler datasets. The full model significantly outperforms all the variants, highlighting the substantial contributions of both the Attention-Aggregation Network and the Gated Multi-scale Fusion Transformer modules to the overall performance. The removal of either of these modules leads to a notable drop in performance metrics.

The Attention-Aggregation Network (AAN) plays a crucial role in addressing color distortion in the low-frequency component of the image, while the Gated Multi-scale Fusion Transformer (GMFT) is essential for refining high-frequency edge features. The combined effect of these modules enables the DocDeshadower to effectively manage shadows across various frequencies and scales, resulting in superior visual quality and shadow removal performance.

These ablation studies demonstrate the importance of each component in the proposed architecture and validate the design choices made in the development of the DocDeshadower method. The results underscore the significance of employing a multi-scale approach and leveraging the strengths of both attention mechanisms and transformer-based architectures for document shadow removal tasks.

\subsection{Limitations and Future Work}
While DocDeshadower demonstrates superior performance in document shadow removal, it is important to acknowledge its limitations and potential areas for future work:
\begin{itemize}
	\item Dataset dependency: The model's performance may vary with different types of document shadows or lighting conditions not well-represented in the training data. Expanding the dataset to include a wider variety of document types, shadow conditions, and lighting scenarios could improve the model's generalization capabilities.
	\item Real-time processing: The current model is not optimized for real-time processing, which could limit its application in scenarios requiring immediate shadow removal. Developing a more lightweight version of the model for real-time applications is another potential area for future work.
\end{itemize}

\section{CONCLUSIONS}
In this study, we propose DocDeshadower, a novel model designed for shadow removal that effectively combines Transformer and Attention mechanisms across multiple frequencies. By addressing both high-frequency textural features and low-frequency color information, our model achieves state-of-the-art performance. The results provide compelling evidence that DocDeshadower's multi-scale approach, coupled with its innovative use of attention mechanisms and transformer-based architectures, enables it to effectively manage shadows and produce high-quality, shadow-free images.






\section*{ACKNOWLEDGMENT}
This work was supported by the Science and Technology Development Fund, Macau SAR, under Grant 0141/2023/RIA2 and 0193/2023/RIA3.

\bibliographystyle{IEEEtran}
\bibliography{ref}

\end{document}